\documentclass[sigconf]{acmart}
\AtBeginDocument{%
  }

\copyrightyear{2025}
\acmYear{2025}
\setcopyright{acmlicensed}
\acmDOI{10.1145/3746027.3755752}
\acmConference[MM '25] {Proceedings of the 33rd ACM International Conference on Multimedia}{October 27--31, 2025}{Dublin, Ireland.}
\acmBooktitle{Proceedings of the 33rd ACM International Conference on Multimedia (MM '25), October 27--31, 2025, Dublin, Ireland}
\acmISBN{979-8-4007-2035-2/2025/10}

\usepackage{tikz}
\usepackage{multirow}
\usepackage{makecell}
\usepackage{subfigure}

\begin{document}

\title{MultiMind: Enhancing Werewolf Agents with Multimodal Reasoning and Theory of Mind}

\author{Zheng Zhang}
\orcid{0009-0001-6227-1987}
\authornote{Equal contribution.}
\affiliation{%
  \institution{The Hong Kong University of Science and Technology (Guangzhou)}
  \city{Guangzhou}
  \country{China}
}
\email{zzhang302@connect.hkust-gz.edu.cn}

\author{Nuoqian Xiao}
\orcid{0009-0000-3392-8778}
\authornotemark[1]
\authornote{This work was done during internship at HKUST(GZ).}
\affiliation{%
  \institution{The Hong Kong University of Science and Technology (Guangzhou)}
  \city{Guangzhou}
  \country{China}
}
\email{xiaonuoqian@sjtu.edu.cn}

\author{Qi Chai}
\orcid{0009-0009-3717-9482}
\affiliation{%
  \institution{The Hong Kong University of Science and Technology (Guangzhou)}
  \city{Guangzhou}
  \country{China}
}
\email{ericedu@stu.xjtu.edu.cn}

\author{Deheng Ye}
\orcid{0000-0002-1754-1837}
\authornotemark[3]
\affiliation{%
 \institution{Tencent}
 \city{Shenzhen}
 \country{China}
}
\email{dericye@tencent.com}

\author{Hao Wang}
\orcid{0000-0002-3086-3128}
\authornote{Corresponding author.}
\affiliation{%
  \institution{The Hong Kong University of Science and Technology (Guangzhou)}
  \city{Guangzhou}
  \country{China}
}
\email{haowang@hkust-gz.edu.cn}

\renewcommand{\shortauthors}{Zheng Zhang, Nuoqian Xiao, Qi Chai, Deheng Ye, \& Hao Wang}

\begin{abstract}
Large Language Model (LLM) agents have demonstrated impressive capabilities in social deduction games (SDGs) like Werewolf, where strategic reasoning and social deception are essential. However, current approaches remain limited to textual information, ignoring crucial multimodal cues such as facial expressions and tone of voice that humans naturally use to communicate. Moreover, existing SDG agents primarily focus on inferring other players' identities without modeling how others perceive themselves or fellow players. To address these limitations, we use One Night Ultimate Werewolf (ONUW) as a testbed and present MultiMind, the first framework integrating multimodal information into SDG agents. MultiMind processes facial expressions and vocal tones alongside verbal content, while employing a Theory of Mind (ToM) model to represent each player's suspicion levels toward others. By combining this ToM model with Monte Carlo Tree Search (MCTS), our agent identifies communication strategies that minimize suspicion directed at itself. Through comprehensive evaluation in both agent-versus-agent simulations and studies with human players, we demonstrate MultiMind's superior performance in gameplay. Our work presents a significant advancement toward LLM agents capable of human-like social reasoning across multimodal domains. Our code is available at \url{https://github.com/CjangCjengh/onuw}.
\end{abstract}



\maketitle

\begin{figure}[ht]
    \centering
    \includegraphics[width=\columnwidth]{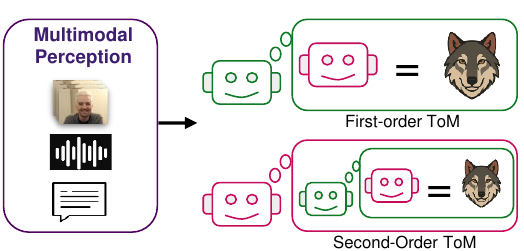}
    \caption{An example of Theory of Mind (ToM). First-order ToM involves a player inferring other players' identities, while second-order ToM extends to reasoning about the individual opinions of others.}
    \label{fig:belief_concept}
\end{figure}

\section{Introduction}

Large Language Model (LLM) agents have made significant progress in simulating human-like decision-making and social behavior across various domains.
Recent research has demonstrated their ability to play games such as StarCraft \citep{ma2024large, shao2024swarmbrainembodiedagentrealtime} and MineCraft \citep{wang2024voyager, 10657187, li2024optimus}. Within this broad research landscape, a promising direction has developed focusing on games that require not just strategic thinking but also sophisticated social reasoning. Social deduction games (SDGs) such as Werewolf \citep{10888525, xu2023exploring, 10.5555/3692070.3694355, wu2024enhancereasoninglargelanguage}, Avalon \citep{light2023from, wang2023avalonsgamethoughtsbattle, shi2023cooperationflyexploringlanguage, lan-etal-2024-llm}, and Jubensha \citep{10.1145/3721121, wu-etal-2024-deciphering} have proven to be challenging testbeds for LLM agent research. These games demand nuanced reasoning and strategic communication based on hidden information, which are skills that more closely resemble human social intelligence.

Despite notable progress in developing LLM agents for SDGs, current approaches exhibit significant limitations. Most existing agents are constrained to processing purely textual information, overlooking crucial multimodal cues such as facial expressions and tone of voice. In real-world settings, these non-verbal signals often reveal underlying intentions, emotional states, and potential deception that complement verbal exchanges. While some research has begun incorporating multimodal data in SDG analysis \citep{lai-etal-2023-werewolf, 10657493}, these approaches primarily focus on retrospective analysis rather than active gameplay participation.


Additionally, current methods typically attempt to infer other players' identities, but fail to model how others perceive the identity of the agent itself or of other players. As shown in Figure \ref{fig:belief_concept}, this layered reasoning structure, known as Theory of Mind (ToM), is the cognitive ability to understand and identify mental states in oneself and others. The lack of systematic ToM modeling capabilities prevents agents from engaging in the multi-level strategic reasoning that human players naturally employ in complex interactions.

In this paper, we focus on One Night Ultimate Werewolf (ONUW) following recent research \citep{lai-etal-2023-werewolf, jin2024learning}, a variant of the traditional Werewolf game. ONUW is particularly suitable for our research as it has available multimodal datasets \citep{lai-etal-2023-werewolf}, such that we may use the multimodal cues to conduct more comprehensive reasoning. Besides, it concentrates gameplay into a single night phase followed by one day phase for discussion and voting, placing emphasis on strategic communication and social reasoning. This creates an ideal testbed for exploring multimodal reasoning in social settings where verbal and non-verbal cues become crucial for communication.

To this end, we present MultiMind, a novel framework that enhances LLM agents for ONUW by integrating multimodal information and ToM reasoning. MultiMind converts facial expressions and tone of voice into textual descriptions that capture emotional signals, enabling our agent to process multimodal information. We further implement ToM reasoning by encoding player statements into discrete action representations, which predicts the belief distributions that represent each player's suspicion levels toward others. Then, we employ Monte Carlo Tree Search (MCTS) to optimize communication strategies based on this ToM model, identifying utterances that minimize suspicion directed at the agent.

The training of the ToM model follows a two-stage approach. We initially use synthetic data from LLM-based agent self-play, then fine-tune on human gameplay data. Through comprehensive evaluation in both agent simulations and studies with human players, we demonstrate substantial improvements in gameplay performance while generating more convincing and strategically sound communication patterns. MultiMind presents a notable step toward LLM agents that can effectively engage in complex social reasoning across multimodal domains.

Our main contributions can be summarized as follows:
\begin{itemize}
    \item To our best knowledge, we are the first to develop a framework that integrates multimodal information into SDG agents, enhancing their realism and strategic capabilities.
    \item We propose a novel approach combining ToM modeling with MCTS to optimize communication strategies, enabling agents to reason about belief states and strategically minimize suspicion directed at themselves.
    \item We demonstrate the effectiveness of our approach through comprehensive evaluation in both agent-versus-agent simulations and studies with human players, showing improvements in gameplay performance.
\end{itemize}

\section{Related Work}

\subsection{Social Deduction Game Agent}

Social deduction games (SDGs) have emerged as valuable testbeds for AI research. Early research \citep{Hirata2016WerewolfGM, Nakamura2016ConstructingAH, Wang2018ApplicationOD, 10.5555/3454287.3454400} on SDG agents employs rule-based or learning-based methods for decision-making. These agents use predefined protocols for communication rather than natural language. With the emergence of LLMs, more sophisticated agents have been developed across various SDGs. For Avalon, \citet{lan-etal-2024-llm} employ system prompts to guide LLM agents in gameplay, while DeepRole \citep{10.5555/3454287.3454400} combine counterfactual regret minimization with value networks trained through self-play. For traditional Werewolf, \citet{xu2023exploring} develop agents that generate diverse action candidates through deductive reasoning and utilize RL policies to optimize strategic gameplay. \citet{wu2024enhancereasoninglargelanguage} further enhance System-2 reasoning abilities by training a Thinker module for complex logical analysis. For One Night Ultimate Werewolf (ONUW), \citet{jin2024learning} propose an RL-instructed language agent framework that determines appropriate discussion tactics.

Despite these advances, current SDG agents rely exclusively on textual information, ignoring non-verbal cues that humans naturally use during social interactions. In real-world settings, facial expressions and tone of voice often reveal underlying intentions, emotional states, and potential deception that complement verbal exchanges. This limitation prevents existing agents from accessing and interpreting the full spectrum of communication signals available in human gameplay.

\subsection{Multimodal Social Interaction}

Research on multimodal social interaction has expanded significantly in recent years, exploring how non-verbal cues complement verbal communication. \citet{Grauman_2022_CVPR} present the Ego4D social benchmark for understanding social attention through video and audio. In the domain of SDGs, \citet{10657493} introduce challenges for modeling fine-grained dynamics using densely aligned language-visual representations. Focusing specifically on ONUW, \citet{lai-etal-2023-werewolf} analyze gameplay recordings to identify behavioral patterns correlating with deception.

However, these works primarily focus on retrospective analysis of social interactions rather than active participation in gameplay. They typically analyze recorded gameplay to extract patterns or identify deception markers after the fact, but do not employ these insights to inform real-time decision-making by autonomous agents. Our work differs by integrating multimodal reasoning directly into an agent's gameplay loop, enabling it to both interpret non-verbal cues from other players and strategically manage its own communication during active gameplay.

\begin{figure*}
    \centering
    \includegraphics[width=\textwidth]{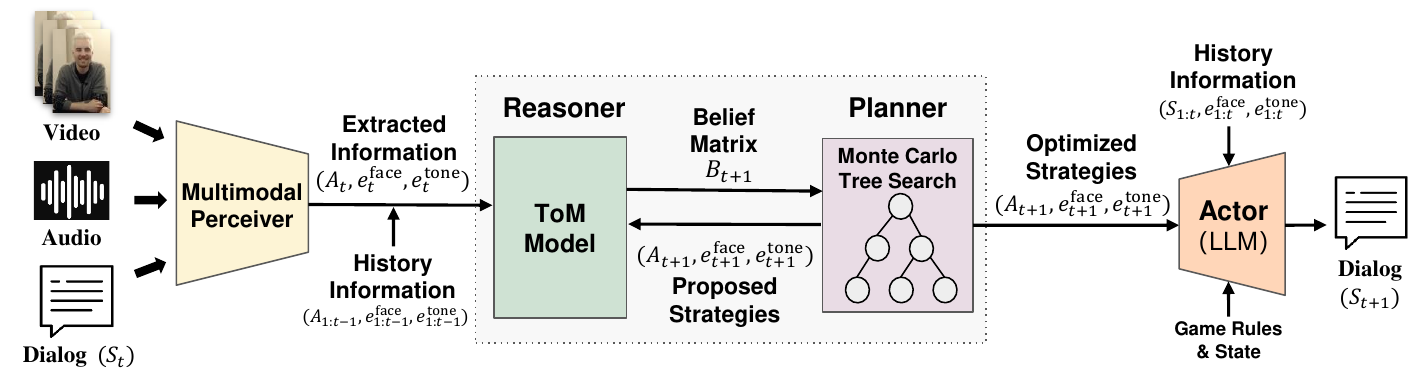}
    \caption{The overall framework of MultiMind. The Perceiver extracts structured information from multimodal inputs. The Reasoner and Planner iteratively optimize communication strategies: the Reasoner predicts each player's belief states using a ToM model, while the Planner employs Monte Carlo Tree Search (MCTS) to explore possible strategies. Finally, the Actor generates natural language statements that implement the optimized strategies.}
    \label{fig:framework}
\end{figure*}

\subsection{Theory of Mind}

Theory of Mind (ToM) refers to the cognitive ability to attribute mental states to oneself and others \citep{Bratman1987-BRAIPA}, allowing agents to reason about beliefs, desires, intentions, and knowledge of other individuals. This capability is fundamental for effective social interaction and has been extensively studied in multi-agent systems.
Traditional computational approaches to ToM have utilized Bayesian methods \citep{BAKER2009329, 8673023, 9209472} to model beliefs and update them based on observed behaviors. More recent work has explored neural network-based approaches \citep{pmlr-v80-rabinowitz18a, bara-etal-2021-mindcraft, zhou-etal-2023-cast} that learn representations of others' mental states directly from data. These approaches have been applied in both observational contexts, where the system infers others' beliefs without interaction \citep{pmlr-v80-rabinowitz18a, Grant2017HowCM}, and interactive settings where agents must coordinate their actions \citep{10.1145/3411764.3445645, qiu-etal-2022-towards}.

In the specific context of SDGs like Werewolf and ONUW, ToM reasoning is essential but has been primarily implemented at a first-order level. First-order ToM involves inferring other players' identities, which most existing work \citep{xu2023exploring, wu2024enhancereasoninglargelanguage, jin2024learning} focuses on. However, second-order ToM reasoning about what others believe about oneself and other players remains largely unexplored in these methods. This limitation is significant because in SDGs, players must not only deduce others' roles but also strategically manage others' beliefs about their own identity.
To address this gap, our work explicitly models second-order ToM by representing each player's suspicion levels toward every other player, including the agent itself. This enables our agent to reason about how its communications affect others' beliefs and to strategically select utterances that manage suspicion directed at itself.

\section{Method}

In this section, we first introduce our framework architecture and its four key components, and then present the data construction and training process of our system.

\subsection{Overview}

As shown in Figure \ref{fig:framework}, Our framework consists of four components: the Perceiver, the Reasoner, the Planner, and the Actor.

\textbf{The Perceiver} extracts essential information from dialogue history and, when playing against humans, processes facial expressions from video and emotional cues from audio.

\textbf{The Reasoner} uses the information provided by the Perceiver to infer second-order beliefs for each player, modeling how players perceive one another's potential identities.

\textbf{The Planner} then leverages these belief states predicted by the Reasoner to determine communication strategies. 

\textbf{The Actor} implements these strategies by formulating coherent statements or making decisions.

These components enable our agent to process multimodal information and engage in sophisticated ToM reasoning, allowing for more human-like social deduction capabilities.

\begin{figure*}[!ht]
    \centering
    \includegraphics[width=\textwidth]{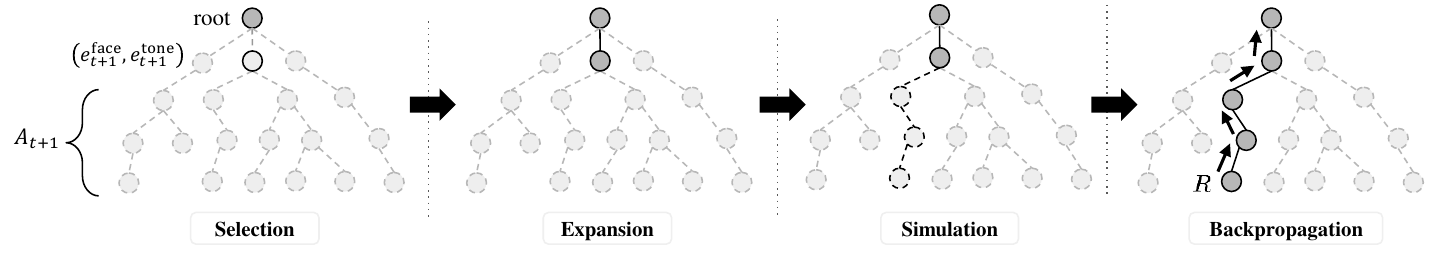}
    \caption{The tree structure of the MCTS algorithm in the Planner. The first level of the tree consists of nodes representing pairs of emotion labels $(e_{t+1}^{\text{face}}, e_{t+1}^{\text{tone}})$. The subsequent levels contain nodes that represent either action triplets $(p_i, a, p_j)$ or “stop” actions, where multiple triplets combine to form the final action sequence $A_{t+1}$. Each iteration of the MCTS involves selection, expansion, and simulation, followed by backpropagation of the reward $R$ from the terminal node back to the root.}
    \label{fig:mcts}
\end{figure*}

\subsection{Perceiver for Information Extraction}
\label{sec:method-perceiver}

The Perceiver extracts structured information from game dialogue and, when playing against humans, processes multimodal signals including facial expressions and vocal tones.

For textual information processing, we employ an LLM to parse each player's statements into structured triplets of (\textit{subject}, \textit{predicate}, \textit{object}). The subject is always the speaker, while the predicate belongs to a finite action space consisting of predefined actions: “support”, “suspect”, and “accuse as \textit{role}” for each possible role. In ONUW, there are five distinct roles, resulting in a predicate space of 2+5=7 possible actions. The object specifies the target of the predicate, which can include the speaker themselves. Formally, we can express this extraction as:

\begin{equation}
A_t = f_{\text{LLM}}(S_t) \subseteq \{(p_i, a, p_j) \mid a \in \mathcal{A}, p_j \in \mathcal{P}\},
\end{equation}
where $S_t$ is the statement made by player $p_i$ at time $t$, $A_t$ represents the set of actions extracted from player $p_i$'s statement, $\mathcal{A}$ is the predicate action space, and $\mathcal{P}$ is the set of players.

When playing against human participants, we enhance information extraction through multimodal processing. Human players are individually seated at computers, where their video and audio are recorded during gameplay. For audio processing, we utilize OSUM \citep{geng2025osum} to classify vocal tones into one of eight emotional categories (happy, sad, neutral, angry, surprise, disgust, fear, or other) and transcribe speech into text. Concurrently, for visual information, we employ Emotion-LLaMA \citep{NEURIPS2024_c7f43ada, 10.1145/3689092.3689404} to classify facial expressions in video frames into the same eight emotional categories. This process can be formalized as:

\begin{equation}
\begin{aligned}
S_t, e_t^{\text{tone}} = f_{\text{audio}}(\text{Audio}_t), \\
e_t^{\text{face}} = f_{\text{visual}}(\text{Video}_t),
\end{aligned}
\end{equation}
where $S_t$ is the transcribed speech, and $e_t^{\text{tone}}$ and $e_t^{\text{face}}$ represent the emotion labels derived from player $p_i$'s audio and video at time $t$, respectively.

The Perceiver integrates these textual actions and emotional signals to provide a comprehensive multimodal representation of each player's communication, which is then passed to the Reasoner for further processing and belief state modeling.

\subsection{Reasoner with ToM modeling}
\label{sec:method-reasoner}

The Reasoner processes the structured information extracted by the Perceiver and employs a specialized Theory of Mind (ToM) model to predict belief distributions representing each player's werewolf suspicions toward others.

Our ToM model adopts a Transformer \citep{NIPS2017_3f5ee243} with causal attention where each input token represents an event in the game timeline, and each corresponding output hidden state is transformed through a linear layer to produce a belief matrix $B$. The causal attention mechanism ensures that the computation of each belief matrix only considers the game history up to the current time point, reflecting the progressive nature of belief formation during gameplay.

For the $k$-th triplet $(p_i, a^{(k)}, p_j^{(k)}) \in A_t$ at time $t$, we encode it as an input token comprising five embeddings:

\begin{equation}
\label{eq:emb_plus}
\begin{aligned}
E_{t,k} = &E_{\text{subj}}(p_i) + E_{\text{pred}}(a^{(k)}) + E_{\text{obj}}(p_j^{(k)}) \\
&+ E_{\text{face}}(e_t^{\text{face}}) + E_{\text{tone}}(e_t^{\text{tone}}),
\end{aligned}
\end{equation}
where $E_{\text{subj}}$, $E_{\text{pred}}$, and $E_{\text{obj}}$ represent embeddings for the subject, predicate, and object respectively. $E_{\text{face}}$ encodes facial emotion labels, and $E_{\text{tone}}$ encodes speech emotion labels. All action triplets from player $p_i$ at time $t$ share the same $e_t^{\text{face}}$ and $e_t^{\text{tone}}$.

Let $N_t = \sum_{\tau=1}^{t} |A_\tau|$ denote the total number of triplets up to time $t$. The complete input sequence $E_{1:N_t}$ comprises the sequential collection of all encoded triplets:

\begin{equation}
\begin{aligned}
E_{1:N_t} = \left[ E_{1,1}, \ldots, E_{1,|A_1|}, \ldots, E_{t,1}, \ldots, E_{t,|A_t|} \right].
\end{aligned}
\end{equation}

A Transformer $\mathcal{D}$ with causal attention processes this sequence to generate hidden states:

\begin{equation}
h_{k} = \mathcal{D}(E_{1:k}), \quad \forall k \in \{1, 2, ..., N_t\}.
\end{equation}

The hidden state $h_{N_t}$ is then processed through a linear layer to generate the belief matrix $B_t$ at time $t$:

\begin{equation}
B_t[i,:] = \text{Softmax}(W_i \cdot h_{N_t} + b_i), \quad \forall i \in \{1, 2, ..., |\mathcal{P}|\},
\end{equation}
where $|\mathcal{P}|$ is the number of players. Each element $B_t[i,j]$ denotes the probability that player $p_i$ believes player $p_j$ is a werewolf.

We can thus encapsulate the belief matrix computation as a function of the history of actions and emotion labels:

\begin{equation}
\label{eq:tom_belief}
B_t = f_{\text{ToM}}(A_{1:t}, e_{1:t}^{\text{face}}, e_{1:t}^{\text{tone}}),
\end{equation}
where $A_{1:t}$ represents all action triplets up to time $t$, and $e_{1:t}^{\text{face}}$ and $e_{1:t}^{\text{tone}}$ represent the corresponding sequences of facial and speech emotion labels, respectively.

This ToM modeling enables our agent to reason how it is perceived by others, providing the Planner with belief state information for optimizing communication strategies.

\subsection{Planner with Monte Carlo Tree Search}

The Planner employs Monte Carlo Tree Search (MCTS) to determine optimal communication strategies based on belief states predicted by the Reasoner. Its primary objective is to generate a sequence of strategic actions $A_{t+1}$ that minimizes the cumulative suspicion directed toward the agent by other players.

To achieve this, the Planner aims to identify a sequence of statements that will most effectively reduce the probability of being suspected as a werewolf. Let $i$ denote the index of our agent in the player set $\mathcal{P}$. Formally, we define the optimization problem as:

\begin{equation}
(A_{t+1}^*, e_{t+1}^{\text{face}*}, e_{t+1}^{\text{tone}*}) = \underset{A_{t+1}, e_{t+1}^{\text{face}}, e_{t+1}^{\text{tone}}}{\arg\min} \sum_{j=1, j \neq i}^{|\mathcal{P}|} B_{t+1}[j, i],
\end{equation}
where $B_{t+1}$ is computed using Equation \eqref{eq:tom_belief}:

\begin{equation}
\label{eq:next_belief}
\begin{aligned}
B_{t+1} &= f_{\text{ToM}}(A_{1:t+1}, e_{1:t+1}^{\text{face}}, e_{1:t+1}^{\text{tone}}) \\
&= f_{\text{ToM}}(A_{1:t} \cup A_{t+1}, e_{1:t}^{\text{face}} \cup \{e_{t+1}^{\text{face}}\}, e_{1:t}^{\text{tone}} \cup \{e_{t+1}^{\text{tone}}\}).
\end{aligned}
\end{equation}

As shown in Figure \ref{fig:mcts}, the MCTS operates on a hierarchical tree structure rooted at the current game state $(A_{1:t}, e_{1:t}^{\text{face}}, e_{1:t}^{\text{tone}})$. The tree's first level represents the selection of emotion labels, where each node corresponds to a specific pair $(e_{t+1}^{\text{face}}, e_{t+1}^{\text{tone}})$. With 8 possible $e_{t+1}^{\text{face}}$ and 8 possible $e_{t+1}^{\text{tone}}$, this level contains 64 nodes.

Below the first level, the tree branches into additional levels that incrementally construct $A_{t+1}$. Each node at these levels represents either an action triplet $(p_i, a, p_j)$ where $p_i$ is always our agent, or a “stop” action that terminates the sequence. In a 5-player game, with $|\mathcal{A}|=7$ (as described in Section \ref{sec:method-perceiver}) and $|\mathcal{P}|=5$, each node can branch into $7 \times 5 + 1 = 36$ possible nodes. The maximum depth ${|A_{t+1}|}_{\text{max}}$ is constrained to 3.

For each node in the tree, we maintain two values: $Q(n)$, the cumulative reward at node $n$, and $N(n)$, the visit count of node $n$. We initialize the root node with $Q(\text{root}) = 0$ and $N(\text{root}) = 0$.

The MCTS proceeds iteratively through four phases:

\textbf{Selection}: The algorithm starts at the root node and recursively traverses the tree. At each node, if the current node has unvisited children or has no children, stop and select this node. Otherwise, move to the child node $n$ that maximizes the Upper Confidence Bound for Trees (UCT):
\begin{equation}
\text{UCT}(n) = \frac{Q(n)}{N(n)} + C \sqrt{\frac{\ln N(\text{parent}(n))}{N(n)}},
\end{equation}
where $\text{parent}(n)$ is the parent node, and $C$ is a constant that balances between exploitation and exploration, conventionally set to 1.414.
If multiple nodes have the same maximum UCT value, one of them is randomly chosen.

\textbf{Expansion}: If the selected node has no children, the algorithm proceeds directly to the backpropagation phase. Otherwise, we randomly select one of its unvisited children, and initialize it with $Q(n) = 0$ and $N(n) = 0$.

\textbf{Simulation}: From the newly initialized node, we randomly traverse down the tree. A terminal state occurs when either $|A_{t+1}|$ reaches 3 or a “stop” action is reached.

\textbf{Backpropagation}: Once we reach a terminal state with a fully constructed action sequence $(A_{t+1}, e_{t+1}^{\text{face}}, e_{t+1}^{\text{tone}})$, we compute $B_{t+1}$ using Equation \eqref{eq:next_belief}, and calculate the reward as:

\begin{equation}
R = -\sum_{j=1, j \neq i}^{|\mathcal{P}|} B_{t+1}[j, i].
\end{equation}

This reward represents the negative sum of suspicion towards our agent $p_i$ from other players.
We then update $Q(n)$ and $N(n)$ for all nodes along the path from the terminal node back to the root:

\begin{equation}
\begin{aligned}
Q(n) &= Q(n) + R, \\
N(n) &= N(n) + 1.
\end{aligned}
\end{equation}

After 500 iterations, the algorithm selects the terminal state $(A_{t+1}, e_{t+1}^{\text{face}}, e_{t+1}^{\text{tone}})$ with the highest $R$. This optimization process enables our agent to identify communication strategies that can minimize other players' suspicion toward itself.

The optimized strategies $(A_{t+1}, e_{t+1}^{\text{face}}, e_{t+1}^{\text{tone}})$ determined by the Planner are then passed to the Actor, which transforms these strategic actions into natural language utterances.

\subsection{Actor for Response Generation}

The Actor transforms the communication strategies determined by the Planner into coherent natural language utterances that can be effectively communicated during gameplay.

For the action sequence $A_{t+1}$, the Actor employs an LLM to generate a cohesive statement that expresses these actions in natural language. This process can be formalized as:

\begin{equation}
\label{eq:gen_resp}
S_{t+1} = g_{\text{LLM}}(\mathcal{R}, G_t, S_{1:t}, e_{1:t}^{\text{face}}, e_{1:t}^{\text{tone}}, A_{t+1}),
\end{equation}
where $S_{t+1}$ is the agent's statement, $\mathcal{R}$ represents the game rules, $G_t$ is the game state at time $t$, $S_{1:t}$ denotes the dialogue history, and $e_{1:t}^{\text{face}}$ and $e_{1:t}^{\text{tone}}$ are the historical facial and speech emotion labels.

For the emotion labels $e_{t+1}^{\text{face}}$ and $e_{t+1}^{\text{tone}}$, the Actor presents them directly as text labels alongside the generated statement. These emotion indicators help human players and other agents interpret the agent's emotional state, adding another dimension to the communication beyond the textual content.

\subsection{Training Process}
\label{sec:training}

In our framework, the only component requiring training is the ToM model in the Reasoner. We employ a two-phase training approach: first utilizing data from agent self-play, followed by fine-tuning with human gameplay data from \citet{lai-etal-2023-werewolf}.

\subsubsection{Self-Play Training}

For the initial training, we generate a dataset through agent self-play in 5-player ONUW games. Each agent is randomly assigned one of three LLM backends: GPT-4o~\citep{openai2024gpt4ocard}, Qwen2.5-14B-Instruct \citep{qwen2025qwen25technicalreport}, or Llama-3.1-8B-Instruct \citep{grattafiori2024llama3herdmodels}. Since the ToM model is not yet available, we modify Equation \eqref{eq:gen_resp} to:

\begin{equation}
S_{t+1}, e_{t+1}^{\text{face}}, e_{t+1}^{\text{tone}} = g_{\text{LLM}}(\mathcal{R}, G_t, S_{1:t}, e_{1:t}^{\text{face}}, e_{1:t}^{\text{tone}}).
\end{equation}

This means that each LLM generates statements and simultaneously selects emotional labels from the predefined sets of 8 facial expressions and 8 vocal tones.

After each agent's speech, we prompt all agents to identify their werewolf suspicions. Let $S_i$ be the set of players suspected by agent $p_i$, and $|S_i|$ be the number of suspected players. We construct the ground truth belief matrix $B_t^{\text{GT}}[i,j]$ as:

\begin{equation}
B_t^{\text{GT}}[i,j] = 
\begin{cases}
\frac{1}{|S_i|} & \text{if } |S_i| > 0 \text{ and } p_j \in S_i, \\
\frac{1}{|\mathcal{P}|} & \text{if } |S_i| = 0, \\
0 & \text{otherwise},
\end{cases}
\end{equation}
where $|\mathcal{P}|$ is the number of all players in the game.

We train the ToM model using a cross-entropy loss between the predicted belief matrices and the ground truth matrices:

\begin{equation}
\mathcal{L} = -\sum_{t=1}^{T} \sum_{i=1}^{|\mathcal{P}|} \sum_{j=1}^{|\mathcal{P}|} B_t^{\text{GT}}[i,j] \log(B_t^{\text{pred}}[i,j]).
\end{equation}

\begin{table*}
\caption{Performance of agents in mixed-agent games. “Participations” counts how many times an agent is selected across 400 games (one agent can be selected multiple times in a single game). “Avg. Votes” denotes the average number of votes received by the agent per game (maximum 5 votes).}
\label{tab:random_agent}
\centering
\begin{tabular}{lccccccc}
\toprule
\multirow{2}{*}{\textbf{Agent}} & \multicolumn{3}{c}{\textbf{In Team Werewolf}} & \multicolumn{3}{c}{\textbf{In Team Village}} & \textbf{Overall} \\
\cmidrule(lr){2-4} \cmidrule(lr){5-7} \cmidrule(lr){8-8}
& Participations & Avg. Votes ($\downarrow$) & Win Rate ($\uparrow$) & Participations & Avg. Votes ($\downarrow$) & Win Rate ($\uparrow$) & Win Rate ($\uparrow$) \\
\midrule
ReAct & 79 & 1.29 & 58.2 & 339 & 1.06 & 33.0 & 37.8 \\
Belief & 66 & 1.44 & 59.1 & 315 & 0.97 & 35.6 & 39.6 \\
LLM-instructed & 82 & 1.20 & 67.1 & 318 & 0.98 & 35.9 & 42.3 \\
RL-instructed & 87 & 1.32 & 57.5 & 290 & \textbf{0.83} & 37.2 & 41.9 \\
\midrule
MultiMind (ours) & 86 & \textbf{1.05} & \textbf{70.9} & 338 & \textbf{0.83} & \textbf{44.4} & \textbf{49.8} \\
\bottomrule
\end{tabular}
\end{table*}

\begin{figure}[ht]
    \centering
    \includegraphics[width=\columnwidth]{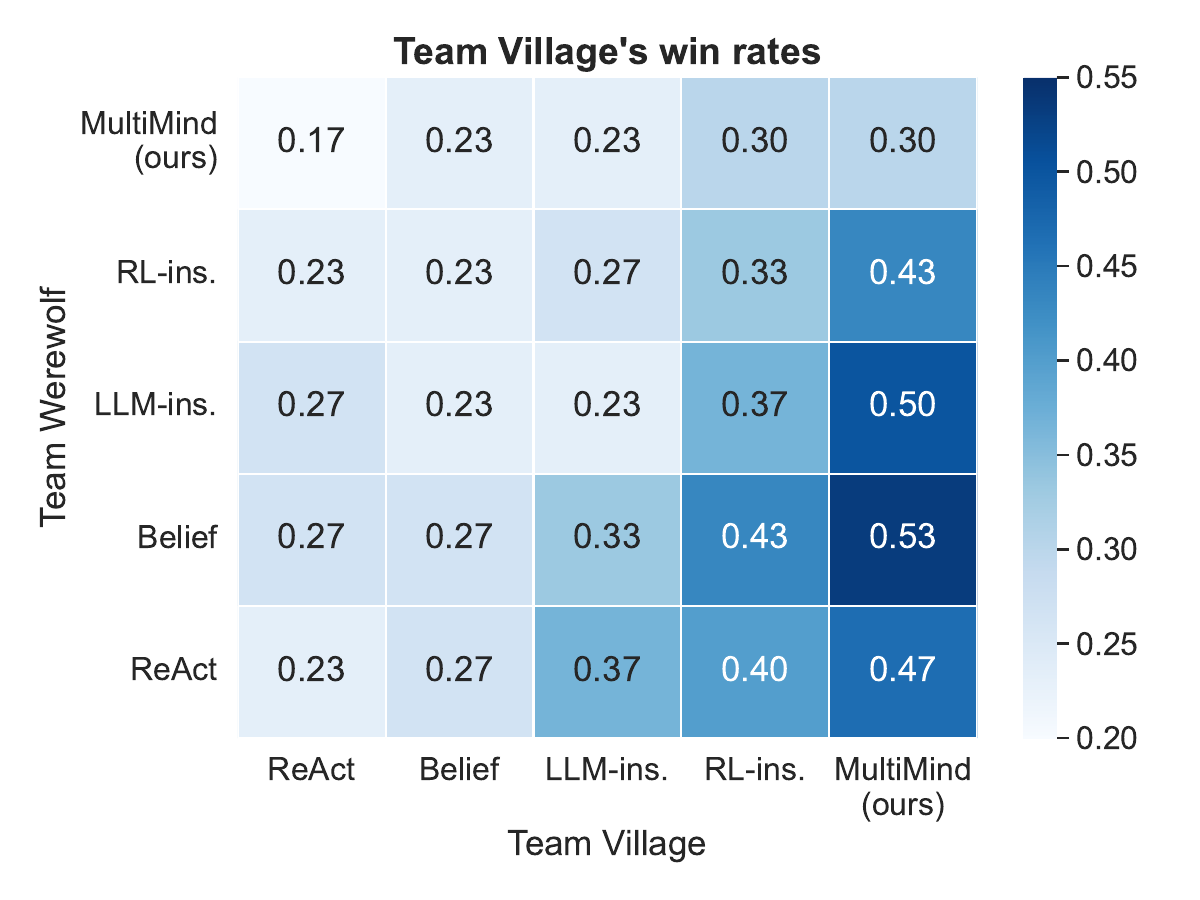}
    \caption{Team Village's win rates. We conduct 30 games for each setting and calculate the win rates.}
    \label{fig:main_wr}
\end{figure}

\subsubsection{Human Data Fine-Tuning}

Following self-play training, we fine-tune our ToM model using the human gameplay dataset from \citet{lai-etal-2023-werewolf}. This dataset comprises 163 videos containing timestamped speech transcriptions and player intention annotations. Each video contains between 1-3 games of 5-player ONUW, totaling 199 complete games.

For each video, we extract individual speech segments using the provided timestamps. To obtain $e_t^{\text{tone}}$, we extract the audio from each speech segment and classify it using OSUM. To obtain $e_t^{\text{face}}$, we leverage Emotion-LLaMA. However, Emotion-LLaMA can only process single-person videos, while the original recordings capture multiple players simultaneously. Fortunately, each player's position remains relatively stable throughout the gameplay as participants are seated. Therefore, we manually crop regions corresponding to each player. These individual player video segments are then processed by Emotion-LLaMA to obtain $e_t^{\text{face}}$.

To construct $B_t^{\text{GT}}$, we utilize the player intention annotations available in the dataset. For segments lacking annotations, we employ GPT-4o by providing it with the complete game record and existing intention annotations, prompting it to infer players' suspicions at specific timestamps.

\begin{table*}
\caption{Comparison of Reasoner variants. Each variant plays against ReAct agent as either Team Werewolf or Team Village. 50 games are conducted for each setting. “Avg. Votes” indicates the average number of votes received by each agent per game (maximum 5 votes). For the ToM model, we use a single 3090 GPU with batch size 1. For Gemini-2.0-Flash, we use 5 threads to simultaneously predict each row of $B_t$.}
\label{tab:ab_reasoner}
\centering
\begin{tabular}{lccccccc}
\toprule
\multirow{2}{*}{\textbf{Reasoner}} & \multirow{2}{*}{\textbf{MCTS iter}} & \multirow{2}{*}{\textbf{\makecell{Planning\\Time (s)}}} & \multicolumn{2}{c}{\textbf{Team Werewolf}} & \multicolumn{2}{c}{\textbf{Team Village}} & \textbf{Overall} \\
\cmidrule(lr){4-5} \cmidrule(lr){6-7} \cmidrule(lr){8-8}
& & & Avg. Votes ($\downarrow$) & Win Rate ($\uparrow$) & Avg. Votes ($\downarrow$) & Win Rate ($\uparrow$) & Win Rate ($\uparrow$) \\
\midrule
ToM model & 1000 & 18.6 & \textbf{0.91} & \textbf{84.0} & \textbf{0.75} & \textbf{50.0} & \textbf{67.0} \\
ToM model & 500 & 7.8 & 0.92 & \textbf{84.0} & 0.77 & 48.0 & 66.0 \\
ToM model & 200 & 4.4 & 1.12 & 76.0 & 0.88 & 26.0 & 51.0 \\
Gemini-2.0-Flash & 200 & 634.2 & 1.09 & 78.0 & 0.89 & 28.0 & 53.0 \\
\bottomrule
\end{tabular}
\end{table*}

\begin{table}[h]
\caption{Comparison of Planner variants. Each variant plays against the ReAct agent as either Team Werewolf or Team Village. For MCTS, we set 500 iterations, which means the ToM model is called 500 times to calculate rewards. For other variants, we traverse 500 nodes, also calling the ToM model 500 times to calculate rewards, and select the path to the highest reward node as the final strategies. We conduct 50 games for each setting.}
\label{tab:ab_planner}
\centering
\begin{tabular}{lccc}
\toprule
\multirow{2}{*}{\textbf{Planner}} & \textbf{\makecell{Team\\Werewolf}} & \textbf{\makecell{Team\\Village}} & \textbf{Overall} \\
& Win Rate ($\uparrow$) & Win Rate ($\uparrow$) & Win Rate ($\uparrow$) \\
\midrule
Random & 76.0 & 38.0 & 57.0 \\
DFS & 68.0 & 20.0 & 44.0 \\
BFS & 80.0 & 42.0 & 61.0 \\
MCTS (Ours) & \textbf{84.0} & \textbf{48.0} & \textbf{66.0} \\
\bottomrule
\end{tabular}
\end{table}

\section{Experiments}
In this section, we first present the results of our agent competing against other baselines, then conduct an ablation study of the ToM model and MCTS, showcase a user study evaluating our agent's performance against human players, and finally discuss the contribution of multimodal cues in ToM modeling.

\subsection{Implementation Details}

\subsubsection{Training Scheme}

Deploying Qwen2.5-14B-Instruct and Llama-3.1-8B-Instruct on a single A800 GPU to complete one game requires an average of 3.5 minutes. We utilized 4 A800 GPUs to generate 5,300 games, which took approximately 77 hours. From this dataset, 5,000 games were allocated to the training set and 300 games to the validation set. We implemented early stopping based on the validation loss, terminating training when the validation loss began to increase. The ToM model, comprising 25.4M parameters (with a hidden size of 512 and 8 decoder-only layers), was trained on a single A800 GPU for 80 epochs with a batch size of 32 and a learning rate of 5e-5, completing in 4.5 hours.

From the 199 games in the human dataset, we allocated 184 games for our training set and 15 games for our validation set. The fine-tuning process for the ToM model was conducted on a single A800 GPU. We employed a batch size of 32 and a learning rate of 5e-5, running for 35 epochs. The entire fine-tuning procedure was completed in 11 minutes.

\subsubsection{Evaluation Setup}

We conduct our experiments using the 5-player ONUW environment\footnote{\url{https://github.com/KylJin/Werewolf}} implemented by \citet{jin2024learning}. This implementation offers two difficulty levels (easy and hard), and we use the hard setting for our experiments. The game includes 1 werewolf, 1 seer, 1 robber, 1 troublemaker, and 1 insomniac. The werewolf belongs to Team Werewolf, while all other roles belong to Team Village. During the daytime phase, all players engage in three rounds of discussion, followed by a voting process. If the werewolf is among those with the highest votes, Team Village wins. Otherwise, Team Werewolf wins.

For comparison, we implement the ReAct \citep{yao2023react} agent, which directly prompts the LLM with raw observations to generate its reasoning and actions. Additionally, we include three agent variants in \citet{jin2024learning}: Belief, LLM-instructed, and RL-instructed.

For agent players, $e_{t}^{\text{face}}$ and $e_{t}^{\text{tone}}$ appear only as text labels generated by the agents themselves. For human players, each human participant is seated individually at a computer where their video and audio are recorded. We then use the methods described in Section \ref{sec:method-perceiver} to extract $S_t$, $e_{1:t}^{\text{face}}$, and $e_{1:t}^{\text{tone}}$.

During the training of our ToM model, we utilized data generated by GPT-4o, Qwen2.5-14B-Instruct, and Llama-3.1-8B-Instruct. Therefore, to demonstrate our generalizability, all experiments use Gemini-2.0-Flash as the backend LLM unless otherwise specified.

When playing against other agents, we use the ToM model trained on self-play data as described in Section \ref{sec:training}, while for games involving human players, we employ the human-data fine-tuned version of the ToM model.

\subsection{Main Results}

Following \citet{jin2024learning}, we evaluate our agent's performance when playing as either Team Village or Team Werewolf against baseline agents. The win rates for Team Village are presented in Figure \ref{fig:main_wr}. The rows of the matrix represent the agent type used by Team Village and the columns represent the agent type used by Team Werewolf. In each game, all players on a given team (Village or Werewolf) use the same agent type.

\begin{table}[h]
\caption{Performance against human players. “Avg. Votes” indicates the average number of votes each agent received per game (maximum 5 votes), and “Avg. Human Votes” represents the average number of votes each agent received from human players per game (maximum 1 vote).}
\label{tab:user_study}
\centering
\begin{tabular}{lccc}
\toprule
\textbf{Agent} & \textbf{\makecell{Avg. Votes\\($\downarrow$)}} & \textbf{\makecell{Avg. Human\\Votes  ($\downarrow$)}} & \textbf{\makecell{Win Rate\\($\uparrow$)}} \\
\midrule
Human & 0.90 & — & 40.0 \\
\midrule
ReAct & 1.18 & 0.32 & 26.2 \\
Belief & 1.16 & 0.28 & 34.8 \\
LLM-instructed & 1.03 & 0.24 & 33.8 \\
RL-instructed & \textbf{0.87} & 0.22 & 40.6 \\
MultiMind & \textbf{0.87} & \textbf{0.19} & \textbf{42.2} \\
\bottomrule
\end{tabular}
\end{table}

As observed across each row, our agent consistently achieves the highest win rates when deployed as Team Village. Similarly, each column reveals that our agent poses significant challenge when playing as Team Werewolf, as evidenced by the low win rates for Team Village when facing our agent.

Notably, our agent's performance boost is more significant when playing as Team Village than as Team Werewolf. We attribute this discrepancy to the experimental setup where all players on a team share identical agent types. This homogeneous team structure particularly amplifies Team Village's advantage by enabling strong coordination through consistent reasoning patterns.

To further evaluate our agent's performance in more realistic scenarios, we conduct 400 games with random agent selection. For each game, the 5 players are randomly selected from our agent and four baseline agents. This setup allows us to assess how each agent performs when integrated into heterogeneous teams, mimicking real-world gameplay with diverse player strategies.

Table \ref{tab:random_agent} presents the results. MultiMind achieves the highest win rate in both Team Werewolf (70.9\%) and Team Village (44.4\%), with the lowest average votes against it across all scenarios. This demonstrates our agent's superior suspicion avoidance, particularly when acting as a werewolf.

\begin{figure}[t]
    \centering
    \includegraphics[width=0.9\columnwidth]{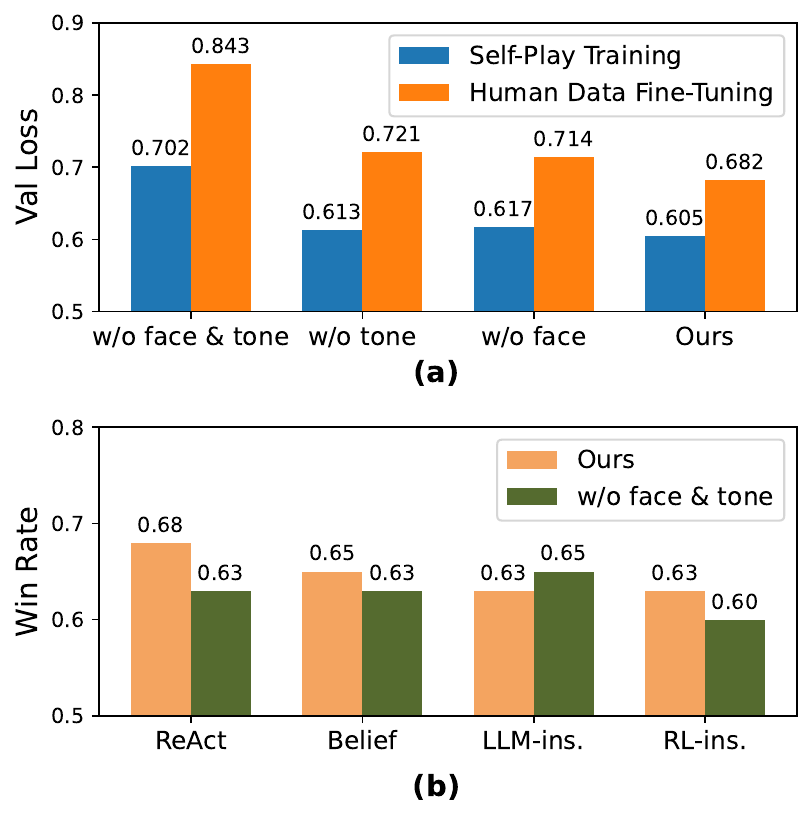}
    \vspace{-0.2cm}
    \caption{Ablation study on multimodal inputs. (a) Comparison of ToM model validation loss. (b) Comparison of win rates. Our agent plays 40 games against each baseline, with 20 games as Team Werewolf and 20 games as Team Village.}
    \label{fig:loss_ablation}
\end{figure}

\subsection{Ablation Study}

To evaluate the contribution of individual components within our framework, we conduct an ablation study focused on the Reasoner and the Planner, which are highly coupled in our architecture.

For the Reasoner, our primary approach uses a lightweight ToM model to predict belief states, allowing the Planner to explore numerous strategies efficiently. As an alternative, we implement a version that directly employs the backend LLM to infer other players' beliefs, modifying Equation \eqref{eq:tom_belief} to:

\begin{equation}
B_t = f_{LLM}(\mathcal{R}, G_t, S_{1:t}, e_{1:t}^{\text{face}}, e_{1:t}^{\text{tone}}),
\end{equation}
where $\mathcal{R}$ represents the game rules, $G_t$ is the game state at time $t$, and $S_{1:t}$ is the complete dialogue history up to time $t$.

Since the LLM-based Reasoner requires significantly more computation time than our ToM model, we change the number of MCTS iterations in the Planner, ensuring a fair comparison. We evaluate all variants against the ReAct agent, with our agent playing as either Team Village or Team Werewolf. The results in Table \ref{tab:ab_reasoner} demonstrate that with the same number of MCTS iterations, our ToM model achieves comparable performance to the LLM-based Reasoner with significantly lower computational cost. This enables us to increase the number of MCTS iterations, spending more time on planning to further improve performance.

The experiments also reveal diminishing returns as MCTS iteration count increases. The performance gap between 200 and 500 iterations is substantial, with overall win rate improving from 51.0\% to 66.0\%. However, increasing from 500 to 1000 iterations yields only negligible improvement.

To demonstrate the contribution of our MCTS Planner, we compare it against alternative planning strategies: random sampling, depth-first search (DFS), and breadth-first search (BFS). Table \ref{tab:ab_planner} presents the results of this comparison. All variants use our ToM model as the Reasoner with identical computational budgets. The MCTS Planner outperforms all alternatives across both Team Village and Team Werewolf scenarios. Notably, DFS performs worse than random sampling due to its tendency to get trapped in suboptimal exploration paths, while BFS achieves relatively good results by providing a more balanced exploration strategy.


\subsection{User Study}

To evaluate our agent's performance in a more realistic setting with human players, we recruited 8 volunteers to participate in our study. After explaining the game rules to these participants, each human player completed 10 games of 5-player ONUW. Each game consisted of 1 human player alongside 4 AI players randomly selected from 5 agents (MultiMind, ReAct, Belief, LLM-instructed, and RL-instructed).

We measured both the win rate of each agent and the number of votes they received. Table \ref{tab:user_study} presents these results. MultiMind achieves the highest win rate while receiving the lowest number of human votes. This aligns with our design objective of suspicion minimization through ToM reasoning.

Notably, while MultiMind still performs well in terms of overall votes received, its advantage is less pronounced compared to our agent-only experiments. We attribute this to the human-data fine-tuned ToM model used in these games. While specializing in human behavior patterns improves deception against humans, it may slightly reduce effectiveness against agent opponents compared to the self-play trained version.


\subsection{Multimodal Cues in ToM Modeling}

To evaluate the impact of multimodal information in ToM modeling, specifically how $e_{t}^{\text{face}}$ and $e_{t}^{\text{tone}}$ influence the prediction of $B_t$, we conduct an ablation experiment. We separately remove the $E_{\text{face}}(e_t^{\text{face}})$ and $E_{\text{tone}}(e_t^{\text{tone}})$ terms from Equation \eqref{eq:emb_plus}, then train the ToM model using the same data and parameters described in Section \ref{sec:training}. To compare performance, we measure the lowest validation loss achieved during both self-play training and human data fine-tuning phases. In both phases, training is stopped when validation loss begins to increase.

As shown in Figure \ref{fig:loss_ablation}(a), both $e_{t}^{\text{face}}$ and $e_{t}^{\text{tone}}$ in Equation \eqref{eq:emb_plus} contribute to the model's ability to predict belief states. The complete model incorporating both modalities achieve lower validation loss compared to ablated variants. This demonstrates that multimodal cues provide valuable additional context for modeling players' mental states in social deduction games. Besides, Figure \ref{fig:loss_ablation}(b) depicts the win rates of our agent against other baselines when multimodal information is not used. While there is a slight decrease in performance, the win rates still exceed 0.6 against all baselines.


\section{Conclusion}

In this paper, we introduced MultiMind, a novel framework that enhances LLM agents for social deduction games by integrating multimodal reasoning and ToM capabilities. Through a combination of ToM reasoning and MCTS planning, MultiMind effectively optimizes communication strategies to minimize suspicion from other players, demonstrating significant improvements in gameplay performance. Our work highlights the potential of multimodal cues in complex social reasoning, paving the way for the development of AI agents capable of sophisticated social interactions.

\begin{acks}
This research is supported by Tencent Rhino-Bird Focused Research Program, National Natural Science Foundation of China (No. 62406267), and the Guangzhou Municipal Science and Technology Project (No. 2025A04J4070).
\end{acks}

\bibliographystyle{ACM-Reference-Format}
\bibliography{sample-base}


\end{document}